\documentclass[review,3p,sort&compress,times,fleqn]{elsarticle}
\usepackage{graphicx}
\usepackage{lineno}
\usepackage{times}
\usepackage{subfig}
\usepackage{amsmath}

\usepackage{hyperref}
\usepackage[hyperref]{xcolor}
\hypersetup{
  colorlinks   = true, 
  urlcolor     = red, 
  linkcolor    = red, 
  citecolor   = red 
}

\journal{Elsevier}

\begin{document}

\begin{frontmatter}

\title{Model-Agnostic Hybrid Numerical Weather Prediction and Machine Learning Paradigm for Solar Forecasting in the Tropics} 

\author[NUS]{Nigel Yuan Yun Ng}
\author[IHPC]{Harish Gopalan}
\author[IHPC]{Venugopalan S. G. Raghavan}
\author[IHPC]{Chin Chun Ooi\corref{mycorrespondingauthor}}
\cortext[mycorrespondingauthor]{1 Fusionopolis Way, \#16-16 Connexis, Singapore 138632, e-mail:ooicc@ihpc.a-star.edu.sg}


\address[NUS]{Department of Physics, Block S12, Science Drive 3, National University of Singapore, Singapore 117551}
\address[IHPC]{Institute of High Performance Computing, 1 Fusionopolis Way, \#16-16 Connexis, Singapore 138632}

\begin{abstract}
Numerical weather prediction (NWP) and machine learning (ML) methods are popular for solar forecasting. However, NWP models have multiple possible physical parameterizations, which requires site-specific NWP optimization. This is further complicated when regional NWP models are used with global climate models with different possible parameterizations. In this study, an alternative approach is proposed and evaluated for four radiation models. Weather Research and Forecasting (WRF) model is run in both global and regional mode to provide an estimate for solar irradiance. This estimate is then post-processed using ML to provide a final prediction. Normalized root-mean-square error from WRF is reduced by up to 40-50\% with this ML error correction model. Results obtained using CAM, GFDL, New Goddard and RRTMG radiation models were comparable after this correction, negating the need for WRF parameterization tuning. Other models incorporating nearby locations and sensor data are also evaluated, with the latter being particularly promising.

\end{abstract}


\end{frontmatter}


\section{Introduction}
The adoption of renewable energy worldwide has increased due to a combination of technological improvements, favorable government policies, and recognition of the impact on climate by fossil fuels \cite{martinez2016value,creutzig2017underestimated,srivastava2013solar,bahadori2013review,zing2008climate}. It is expected that this adoption will accelerate in the coming years, with more than 50\% of the world's energy needs eventually expected to come from renewable sources. In particular, innovations in material sciences and widespread adoption has decreased the cost of photovoltaic (PV) panels \cite{song2017technoeconomic,el2011review,parida2011review,jelle2012building}, thereby greatly enhancing the viability of solar energy. For example, countries such as Singapore have set a target of attaining 2 gigawatt-peak (GWp) from solar energy by 2030, with ambitions to further decrease existing dependence on fossil fuels \cite{ema2030}. 

Nonetheless, a perennial challenge to the adoption of solar or wind energy is the issue of intermittency. Conventional power plants are able to control their energy output to match the requisite power demand. However, solar irradiance does not remain constant from day to day, and is subject to fluctuations in external variables such as cloud and aerosol patterns \cite{yin2020impacts,saarinen2015power,sovacool2009intermittency,sayeef2012solar,traube2012mitigation}. This intermittency in the production of power negatively affects operational reliability. To emphasize the severity of the problem and to avoid disruptions in the power grid, some countries worldwide require day-ahead forecasting of the power production, and impose steep fines on solar and wind energy producers when the power production does not match the forecast \cite{yang2019operational,yang2021operational}. 

Solar irradiance forecasting is consequently receiving widespread attention around the world to improve the reliability of operation, inspiring efforts such as the Global Energy Forecasting Competition in 2014 \cite{hong2016}. A plethora of irradiance forecasting techniques exist in the literature. Some of the commonly used methods are: statistical models \cite{voyant2017machine,zendehboudi2018application,chaabene2008neuro,sivaneasan2017solar,atique2019forecasting}, ground-based sky or satellite image analysis \cite{dong2014satellite,lorenz2004short,peng20153d,caldas2019very,chow2011intra}, numerical weather prediction \cite{mathiesen2011evaluation,jimenez2016wrf,perez2013comparison,perez2010validation,lopes2018short} and hybrid methods using an ensemble of models \cite{yang2019post,diagne2014post,che2019novel}. A comprehensive review of the different forecasting methods can be found in Refs. \cite{diagne2013review,ren2015ensemble,law2014direct,schultz2021can}.  

The choice of method depends quite significantly on the required forecasting time horizon. In particular, NWP-based methods generally out-perform other methods for forecasts beyond 4-6 hours. However, there are several limitations to the use of NWP-based methods. First, the spatial resolution of open-source NWP simulations is on the order of several kilometers or higher. Second, ground-based corrections are only applied a few times a day. For example, Global Forecast System (GFS) outputs are available only every six hours. Commercial solutions such as IBM-GRAF provide more frequent corrections but are expensive. These limitations are particularly important in the tropics \cite{aryaputera2015day,verbois2018solar,huva2020comparisons,koh2009improved} as NWP codes such as the Weather Research and Forecasting Model (WRF) \cite{skamarock2019description} are widely tested primarily for use in the mid-latitudes. In particular, much effort is typically devoted to exploring the impact of different physical model parameterizations to improve the accuracy of these predictions, including for the tropics \cite{zempila2016evaluation,sun2019validation,chen2017impacts,madala2019customization,madala2020numerical}.

Hybrid methods to post-process NWP output predictions to minimize systemic bias and account for local effects are a potential solution \cite{che2019novel}. The hybrid method is performed in two steps. First, NWP simulations are performed for the area and period of interest. Next, statistical learning methods are applied to learn appropriate non-linear corrections to the NWP predictions, based on discrepancies between historical NWP output and actual measurement data. While the hybrid method is seemingly straightforward, several details must be attended to in implementation. For example, boundary conditions can be applied to the simulations at 00, 06, 12, and 18 Coordinated Universal Time (UTC) for the entire simulation duration for hindcasting. However, in day-head forecasting, such data is unavailable and must not be included in our evaluations. 

In this study, an alternative method of performing hybrid forecasting is proposed and evaluated for the tropics. WRF is run both as a global and regional model before statistical methods are used to further improve the WRF-derived solar irradiance outputs. Initial comparisons with 3 years of met mast data for the month of January demonstrate that the proposed hybrid method provides better performance than NWP alone. The accuracy can also be improved using the previous day or hour's solar irradiance as measured at the location of interest (e.g. a PV farm). The proposed method can also be extended for wind farms by training a corresponding statistical model for wind speed and direction as produced by a NWP or WRF model. More critically, we demonstrate that this approach can circumvent the need for lengthy evaluation of different WRF model parameterizations, while providing improved forecasting performance.

The paper is organized as follows. First, the numerical weather prediction and the machine learning algorithms are described in Section \ref{Section:2}. Section \ref{Section:3} presents the results. Normalized root-mean-square error is presented to quantify the systemic bias in the forecasting and their improvement through this hybrid methodology. A few possible endogeneous models are also compared. Section \ref{Section:4} comprises a brief discussion of the current study's findings and possible extensions, while Section \ref{Section:5} ends with a brief summary of the work.

\section{Methods} \label{Section:2}

\subsection{Numerical Weather Prediction}
The NWP simulations are performed using the WRF model \cite{skamarock2019description}. WRF is a mesoscale weather prediction system developed jointly by National Center for Atmospheric Research (NCAR), the National Oceanic and Atmospheric Administration (NOAA), and the National Centers for Environmental Prediction (NCEP). WRF was initially designed to run as a regional climate model. However, modifications to the code have been available since v3.0.1 to run as a global or regional model \cite{richardson2007planetwrf}.  Additional enhancements for solar energy applications (WRF-Solar) \cite{jimenez2016wrf} have been added and made available since v4.2. These capabilities make it possible to run WRF to provide boundary data for day-ahead forecasting of solar irradiance. However, most NWP studies do not utilize the global mode of WRF to generate boundary conditions for the down-scaled forecasts as it adds a significant amount of computational overhead. As using WRF as a regional model alone limits our ability to correct any systemic bias when information is propagated down from the global scale, this current study performs down-sampling starting from the use of WRF as a global model.

The current study uses WRF v4.0.3 to run the simulations. As the solar enhancements were not yet available in this version, the codebase was modified to include the  WRF-Solar enhancements, although this is not expected to impact the findings of this work. The physics schemes employed in this study are listed in Table \ref{Tab:1}. While there are several physics parameterization options available in WRF, the different simulation groups (M1-M4) use the same parameterization for all physics schemes except that for long-wave and short-wave radiations. In addition to the radiation schemes, the choice of microphysics schemes may also strongly impact the accuracy of solar forecasting. However, as this work is a proof-of-concept demonstration, only the radiation models were varied. 

\begin{figure}[htbp]
\begin{center}
\centerline{\includegraphics[width=0.5\textwidth]{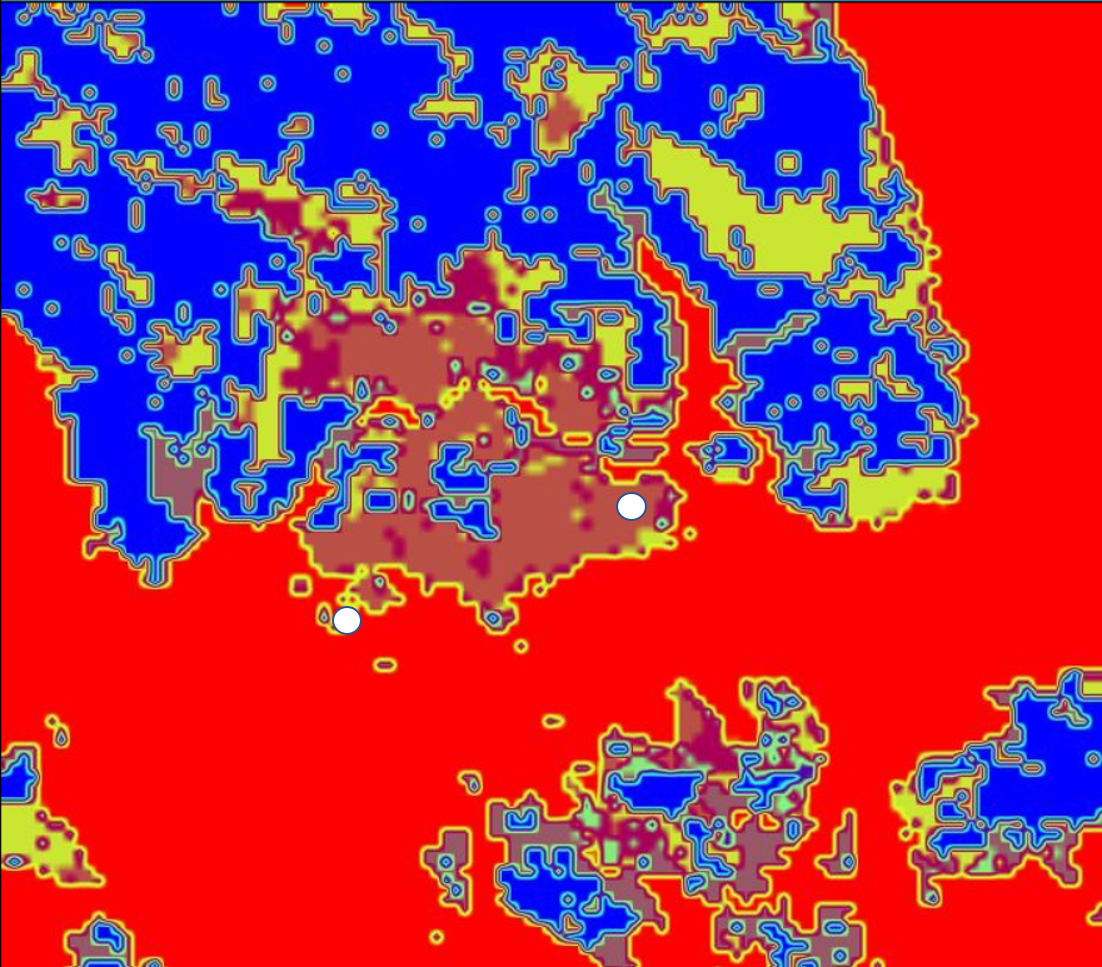}}
\caption{Domain 5 layout color-coded (red represents water body) with the \textit{LU\_INDEX} variable. The two white circle markers represent the locations of the met mast. The top-right one corresponds to the Changi met mast while the bottom one corresponds to the Semakau island met mast.}
\label{Fig:1}
\end{center}
\end{figure}

The simulations were performed on five-nested grids with a horizontal grid resolution of 81, 27, 9, 3, and 1 km. Domain 1 used 487 X 244 cells and was run in global mode. The output from domains 1-4 was saved every 2 hours to provide the boundary conditions for the next set of down-scaled simulations. The microphysics and scalar variables were also used for driving the simulations by setting \textit{ have\_bcs\_moist} and \textit{ have\_bcs\_scalar} to true in the WRF \textit{namelist.input} file. Domains 2-5 used 121 X 121 cells and covered the South-east Asian region, as Singapore was the subject of study for this work. The land usage information for domain 5 is shown in Figure \ref{Fig:1}. All the simulations used 48 vertical levels. The cumulus cloud schemes are usually required only in larger domains as the microphysics models can explicitly resolve the clouds in smaller domains. This is a common method used in many studies \cite{aryaputera2015day,verbois2018solar,huva2020comparisons}. However, there is no clear indication on when to turn off the cumulus cloud schemes. As a result, these schemes were not turned off in any of the domains in the current study.

\renewcommand\arraystretch{1.25}
\begin{table}[htbp]
\centering
\begin{tabular}{|l|c|}
\hline
Flow Physics & Model \\
\hline
Short-Wave radiation & CAM [M1], RRTMG [M2], GFDL [M3] and New Goddard [M4] \\ 
\hline 
Long-Wave radiation & CAM [M1], RRTMG [M2], GFDL [M3] and New Goddard [M4] \\
\hline 
Cumulus & Grell-Freitas scheme \cite{grell2014scale}\\
\hline 
Microphysics & WRF Single-Moment 6-Class microphysics scheme \cite{Hong2006TheWS} \\
\hline
Planetary boundary layer & Yonsei University (YSU) scheme \cite{hong2006new}\\
\hline 
Surface layer & MM5 \cite{jimenez2012revised}\\
\hline
Land surface model & NOAH-MP \cite{niu2011community,yang2011community}\\ 
\hline 
\end{tabular}
\caption{Physics schemes employed in the WRF simulations. Long-wave and Short-wave radiation schemes \cite{collins2004description,iacono2008radiative,fels1981efficient,newgoddard1,newgoddard2} are the only change in parameters between models M1-M4.}
\label{Tab:1}
\end{table}
\renewcommand\arraystretch{1.0}

\subsection{Machine Learning}

Actual meteorological data for two different locations in Singapore, Semakau island and Changi, were acquired at hourly intervals over the period of 1st - 31st January 2014, 1st - 31st January 2015 and 1st - 31st January 2016. January was chosen as the time period for study as it corresponds to Singapore's monsoon period, and is thus expected to be a more challenging season for forecasting due to potential variability in cloud cover. 

For each location, temperature (T), relative humidity (RH), global horizontal irradiance (GHI), and diffuse horizontal irradiance (DHI) were acquired by Meteorological Service Singapore. In addition, direct irradiance (DI) is also a commonly used metric for solar power forecasting. Hence, we calculated the values for DI based on Eq.~\ref{eq:di} and included DI as one of the predictive features. As GHI is the quantity of interest most directly relevant to solar energy production, all our machine learning models in this work were trained to predict GHI based on the WRF output. The two locations were individually trained and tested with our proposed forecasting methods, and results were reported to verify that the findings in this work are reproducible across different locations. 

\begin{equation} \label{eq:di}
   DI = GHI - DHI
\end{equation} 

Tree-based regression methods are a long-established class of machine learning methods that have shown great promise across a wide range of problems. In particular, random forests and gradient boosting are two particular implementations that have consistently yielded good results across different domains \cite{chen2016xgboost, ooi2021modeling,pal2005random,boulesteix2012overview,zhang2018data}. Hence, in this work, two different packages (xgBoost and the Random Forest implementation in scikit-learn) are chosen and used for training and prediction. These models are chosen specifically to represent two common classes of tree-based models, namely bagging (Random Forest) and boosting (xgBoost), and their performance is compared in this work. In addition, a simple k-nearest neighbors (K-NN) regression model is included for comparison (implementation from the scikit-learn library). K-NN models are fast and easy to train, with very few hyper-parameters that need to be tuned, and have similarly been shown to perform fairly well across many problems \cite{al2013multivariate,fan2019application}. Variations of these models have also been shown to work well for solar forecasting in particular \cite{zhong2018xgbfemf, zhang2015gefcom2014,nagy2016gefcom2014,huang2016semi}. 

Most machine learning methods, including random forest and gradient boosting, have model hyper-parameters that need to be optimized. Hence, a randomized grid search with a 5-fold cross-validation split was used to optimize the hyper-parameters for all the models presented in this work. The full data set was split into a training data set (80\% of data) and a test data set (20\% of data) prior to training and hyper-parameter optimization. The test data could then be used to fairly evaluate the trained models without any bias from the training and hyper-parameter optimization. 

The loss function presented in this work is the normalized root-mean-square error (NRMSE), as defined by Eq.~\ref{eq:nrmse}. In addition, as the night irradiance is always zero, only hourly values between 8am and 7pm daily are considered, as these correspond to the typical daylight hours in an equatorial climate like Singapore.

\begin{equation} \label{eq:nrmse}
   NRMSE = \frac{\sqrt{\frac{1}{n}\sum (GHI_{pred} - GHI_{true})^2}} {\frac{1}{n} \sum (GHI_{true})}
\end{equation} 

where $GHI_{pred}$ and $GHI_{true}$ are the predicted and actual quantities of interest.

Outputs for T, RH, GHI and DHI were extracted from the WRF simulations for each day, and used as inputs for the machine learning models, along with corresponding met mast data. Essentially, the machine learning models are being used to correct for systemic predictive errors in the WRF output. In total, we evaluated the use of four different radiation models and combinations of meteorological data for the location of interest as features for the machine learning models. 

As a baseline comparison, we also calculated the predictive performance for GHI if one were to use the WRF outputs directly, or if one were to use a simple day-ahead persistence model, e.g. the predicted GHI value at 12 noon tomorrow is the GHI value at 12 noon today.

\section{Results} \label{Section:3}

\subsection{Baseline Models}

As a baseline for comparison, the NRMSE for day-ahead predictions of GHI was calculated for 2 separate sets of models. 

For the first set of models, the average NRMSE was calculated based on the WRF-obtained predictions for 4 different radiation models. 

\begin{subequations}
\label{eq:WRF}
\begin{align}
   GHI_{pred} (t) = GHI_{wrf}^{M1} (t) \label{eq:WRF1} \\
   GHI_{pred} (t) = GHI_{wrf}^{M2} (t) \label{eq:WRF2} \\
   GHI_{pred} (t) = GHI_{wrf}^{M3} (t) \label{eq:WRF3} \\
   GHI_{pred} (t) = GHI_{wrf}^{M4} (t) \label{eq:WRF4} 
\end{align}
\end{subequations}

The superscripts, M1 to M4, in Eq.~\ref{eq:WRF} refer to the same four models listed in Table ~\ref{Tab:1}.

Next, the average NRMSE was calculated based on a simple persistence model as per Eq.~\ref{eq:persistence}, whereby the GHI was predicted to be identical to the meteorological station's measured GHI value from 24 hours before. As a comparison, we also computed the NRMSE for a 1-hr ahead persistence model. This is expected to yield a much more accurate prediction, although this method is clearly not possible for day-ahead forecasting.

\begin{subequations} 
\label{eq:persistence}
\begin{align}
   &GHI_{pred} (t) = GHI_{met} (t-24) \\
   &GHI_{pred} (t) = GHI_{met} (t-1) 
\end{align}
\end{subequations} 

As per this notation, the ground truth to which all predictions are compared are the met mast data corresponding to $GHI_{met} (t)$. 

\begin{figure}[htbp]
\begin{center}
\centerline{\includegraphics[width=0.5\textwidth]{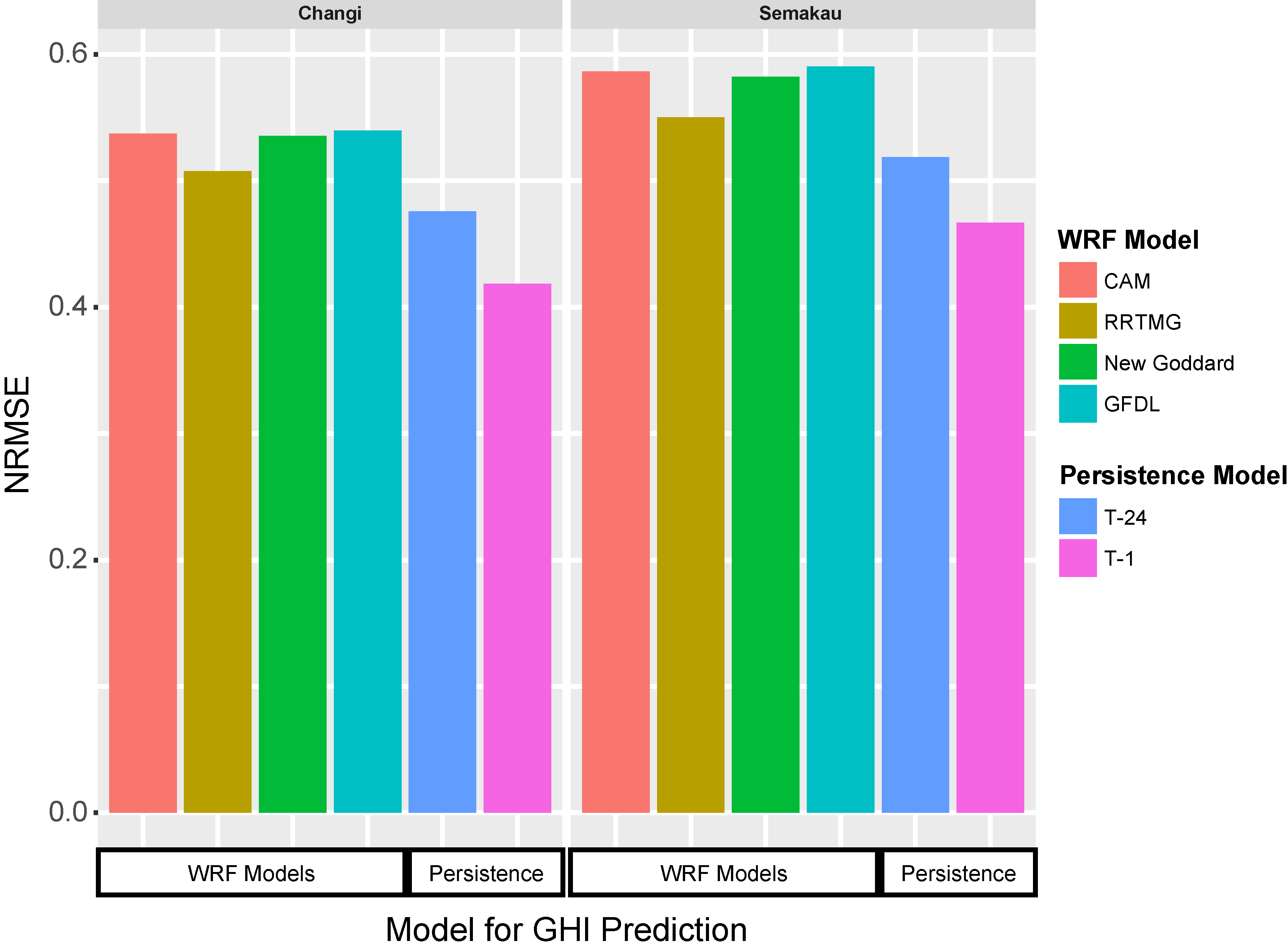}}
\caption{Plots of the normalized root-mean square error (NRMSE) from different WRF radiation models and persistence models}
\label{fig:baseline}
\end{center}
\end{figure}

The results are plotted in Fig.~\ref{fig:baseline} for both Changi and Semakau. As the two locations are quite different, with Semakau in particular being an offshore island with frequent land reclamation projects, the baseline errors are slightly different. This can be attributed in part to the fact that the individual predictions will vary with factors such as different local land-use information. 

In general, the RRTMG radiation models appear to be slightly more suited to Singapore's tropical climate, with slightly lower NRMSE. The persistence model is actually slightly superior to the use of base WRF models for prediction, with the hour-ahead forecasting exhibiting a much lower NRMSE. This is also consistent with typical reports in literature showing that statistical and persistence models work better for short time horizon forecasting, while numerical weather prediction methods gradually become more meaningful for longer time horizons.

\subsection{WRF-ML Models}

Three machine learning methods: i) K-nearest neighbors (KNN) regression, ii) random forest (RF) and iii) xgBoost (XGB), were then used to predict the GHI based on the WRF outputs from each of the four different radiation models as per Eq.~\ref{eq:base_model}.

\begin{align} \label{eq:base_model}
   GHI_{pred} (t) = f \begin{pmatrix} & & &t & & \\
	                        &T_{wrf} (t) & RH_{wrf}(t) & GHI_{wrf} (t) & DHI_{wrf} (t) & DI_{wrf} (t) \end{pmatrix}
\end{align} 

\begin{figure}[htbp]
\begin{center}
\centerline{\includegraphics[width=0.5\textwidth]{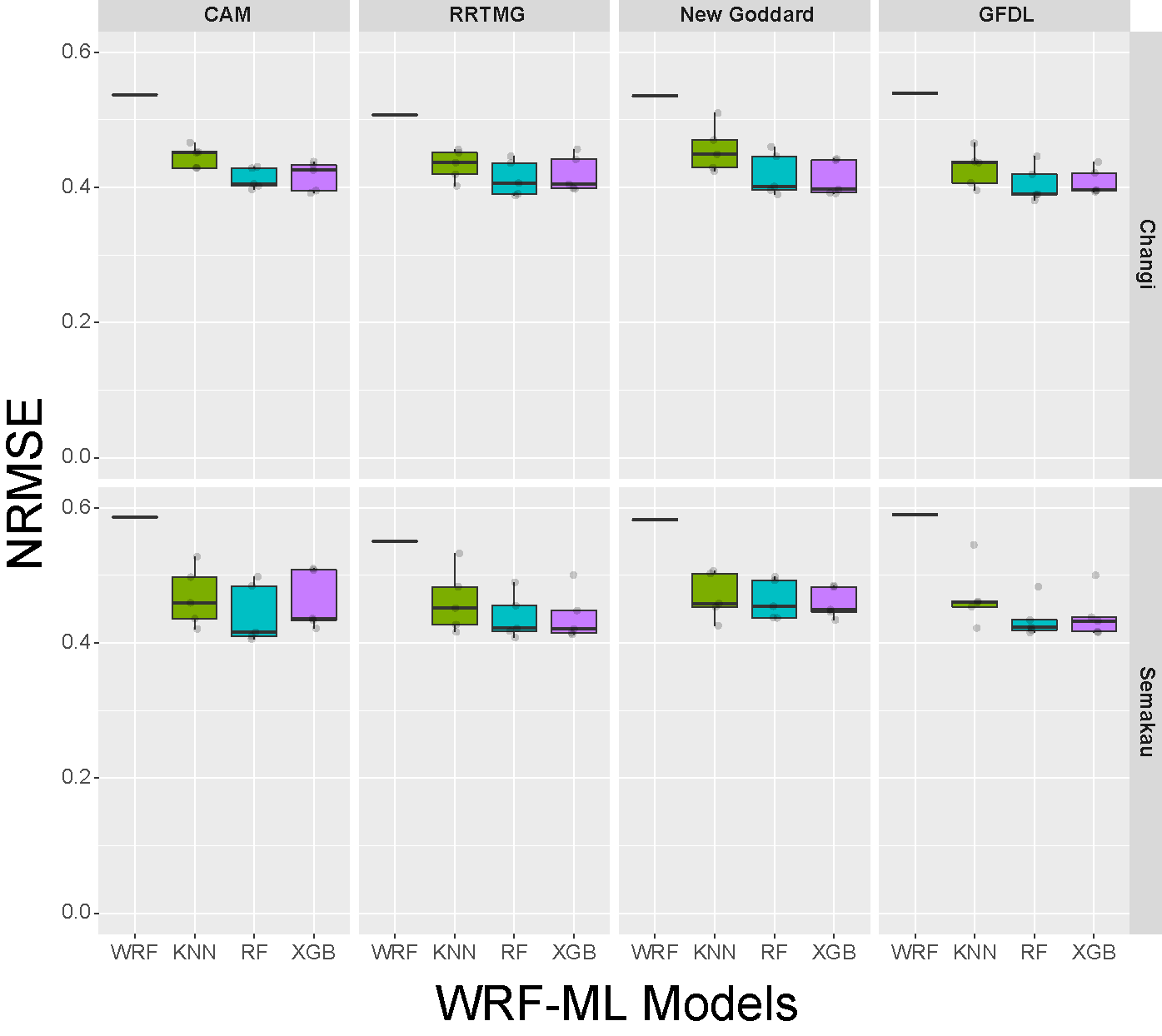}}
\caption{Plots of the normalized root-mean square error (NRMSE) for combinations of different WRF radiation models and ML models}
\label{fig:WRF-ML-Base}
\end{center}
\end{figure}

The function \emph{f} in Eq.~\ref{eq:base_model} can represent either a KNN model, gradient boosting model or random forest model. The NRMSEs are calculated for models constructed from the outputs of the four WRF radiation models and compared in Fig.~\ref{fig:WRF-ML-Base}. In the majority of cases, the gradient boosting and random forest models perform slightly better than the K-NN models. The results show that the use of ML models as data-driven error corrections to the WRF outputs can significantly improve NRMSEs by between 16 and 26 \% across all 4 radiation models, with the largest improvement typically observed for the GFDL radiation model. Interestingly, while RRTMG performed best for the base WRF predictions, the uneven improvement from the ML error correction on each of the four WRF radiation models resulted in all radiation models now having similar predictive performance.

\subsection{Utilizing Neighboring Locations}

As the prediction of GHI is heavily dependent on accurate prediction of cloud cover and movement, knowledge of neighboring locations' GHI might provide information regarding the presence of significant cloud cover. Hence, a set of models were also trained to evaluate the benefits of including WRF outputs for three additional neighboring locations as predictive features in the models. The three locations selected for Changi are approximately 10 km away, and are to the North and Northwest of Changi. Similarly, the 3 locations selected for Semakau Island are approximately 10 km away, and to the North, Northeast and Northwest of the location. The distances are much larger than the finest grid size resolution of 500 m used in the WRF simulations, hence the inclusion of more locations is expected to provide useful additional information. These locations were chosen as they are the 3 nearest met station locations relative to Changi and Semakau island.

\begin{align} \label{eq:nn_model}
   GHI_{pred}^{X_{0}} (t) = f \begin{pmatrix} & & &t & & \\
	                        &T_{wrf}^{X_{0}} (t) & RH_{wrf}^{X_{0}} (t) & GHI_{wrf}^{X_{0}} (t) & DHI_{wrf}^{X_{0}} (t) & DI_{wrf}^{X_{0}} (t) \\
													&T_{wrf}^{X_{1}} (t) & RH_{wrf}^{X_{1}} (t) & GHI_{wrf}^{X_{1}} (t) & DHI_{wrf}^{X_{1}} (t) & DI_{wrf}^{X_{1}} (t) \\
													&T_{wrf}^{X_{2}} (t) & RH_{wrf}^{X_{2}} (t) & GHI_{wrf}^{X_{2}} (t) & DHI_{wrf}^{X_{2}} (t) & DI_{wrf}^{X_{2}} (t) \\
												  &T_{wrf}^{X_{3}} (t) & RH_{wrf}^{X_{3}} (t) & GHI_{wrf}^{X_{3}} (t) & DHI_{wrf}^{X_{3}} (t) & DI_{wrf}^{X_{3}} (t) \end{pmatrix}
\end{align}

The superscripts, $X_{0}$ to $X_{3}$, represent the different spatial locations used in the model, with $X_{0}$ in particular representing either Changi or Semakau. 

\begin{figure}[htbp]
\begin{center}
\centerline{\includegraphics[width=0.5\textwidth]{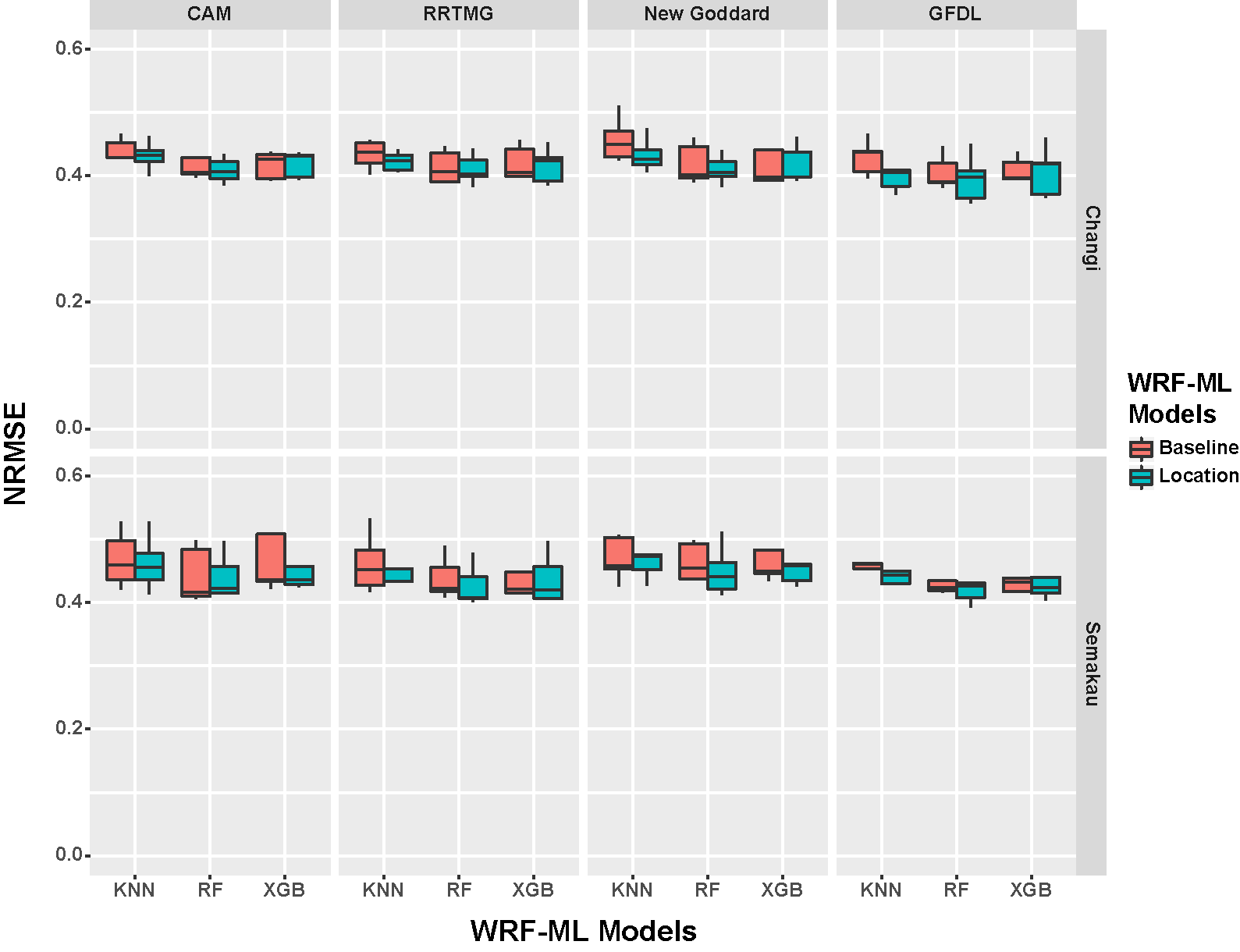}}
\caption{Plots of the normalized root-mean square error (NRMSE) for WRF-ML models with and without the incorporation of neighboring location information}
\label{fig:locations}
\end{center}
\end{figure}

As per Fig.~\ref{fig:locations}, the inclusion of other WRF output information such as irradiance at nearby locations did not improve the predictive performance of the trained models, suggesting that the errors in prediction from the WRF model are probably consistent across these spatial scales. 

\subsection{Utilizing Historical Data from Meteorological Station}

As WRF is essentially a transient physics simulation, a model's deviation from the physical world early on is expected to be correlated to increased divergence at a later time point in the same simulation. Hence, a set of models are also tested, where actual met mast data for the location of interest are provided as features for the predictive model. Hence, the models described in Eq.~\ref{eq:t_model_1} and Eq.~\ref{eq:t_model_24} were tested.

\begin{align} \label{eq:t_model_1}
   GHI_{pred} (t) = f \begin{pmatrix} & & &t & & \\
	                        &T_{wrf} (t) & RH_{wrf} (t) & GHI_{wrf} (t) & DHI_{wrf} (t) & DI_{wrf} (t) \\
													&T_{met} (t - 1) & RH_{met} (t-1) & GHI_{met} (t - 1) & DHI_{met} (t - 1) & DI_{met} (t - 1) \end{pmatrix}
\end{align}

\begin{align} \label{eq:t_model_24}
   GHI_{pred} (t) = f \begin{pmatrix} & & &t & & \\
	                        &T_{wrf} (t) & RH_{wrf} (t) & GHI_{wrf} (t) & DHI_{wrf} (t) & DI_{wrf} (t) \\
													&T_{met} (t - 24) & RH_{met} (t-24) & GHI_{met} (t - 24) & DHI_{met} (t - 24) & DI_{met} (t - 24) \end{pmatrix}
\end{align}

\begin{figure}[htbp]
\begin{center}
\centerline{\includegraphics[width=0.5\textwidth]{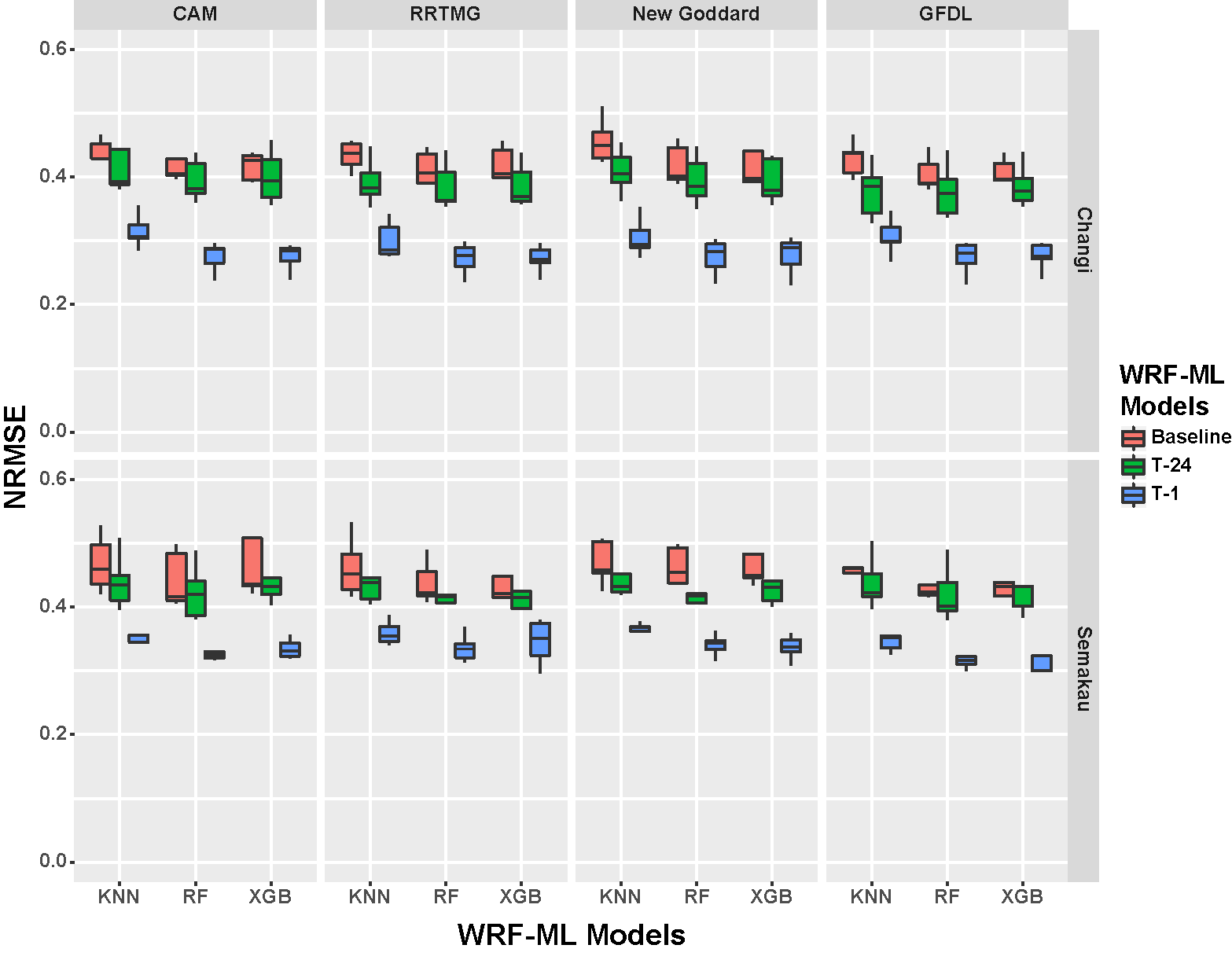}}
\caption{Plots of the normalized root-mean square error (NRMSE) for WRF-ML models with and without the incorporation of previous meteorological information}
\label{fig:time-models}
\end{center}
\end{figure}

As shown in Fig.~\ref{fig:time-models}, the use of actual meteorological readings that approach the required time of prediction can greatly improve the predictive performance of the WRF-ML models. The use of (t-1) data is especially helpful to improving the predictive performance, with a reduction in NRMSE of approximately 40 to 50 \%. Nonetheless, it is important to note that the use of (t-1) meteorological data is not possible when doing day-ahead forecasting. The impact of including (t-24) data is not as significant, with a reduction in NRMSE of approximately 20 to 30 \% relative to the base WRF models, which is an improvement of approximately 4 to 5 \% over the baseline WRF-ML models presented in Eq.~\ref{eq:base_model}. Nonetheless, this does demonstrate the utility of further improved hybrid models comprising WRF-ML-Data schemes where meteorological data obtained at different times could improve forecasting accuracy over different time horizons.

In addition, the gradient boosting and random forest models out-perform the K-NN models in our experiments, with a consistent improvement observed for both Changi and Semakau.

\subsection{Ensemble Model}

As the four radiation models are individually weak predictors for the actual day-ahead GHI value, a model was tested that used the outputs of the four radiation models as features in the model, as indicated in Eq.~\ref{eq:four_models_base}. This is also inspired by prior work showing potential improvements in predictive performance by ensembling machine learning models \cite{alhajj2018forecasting,ahmed2016ensemble}, although the ensemble here is of radiation models as opposed to the ML models.

\begin{align} \label{eq:four_models_base}
   GHI_{pred} (t) = f \begin{pmatrix} & & &t & & \\
	                        &T_{wrf}^{M1} (t) & RH_{wrf}^{M1} (t) & GHI_{wrf}^{M1} (t) & DHI_{wrf}^{M1} (t) & DI_{wrf}^{M1} (t)\\
													&T_{wrf}^{M2} (t) & RH_{wrf}^{M2} (t) & GHI_{wrf}^{M2} (t) & DHI_{wrf}^{M2} (t) & DI_{wrf}^{M2} (t)\\
													&T_{wrf}^{M3} (t) & RH_{wrf}^{M3} (t) & GHI_{wrf}^{M3} (t) & DHI_{wrf}^{M3} (t) & DI_{wrf}^{M3} (t)\\
													&T_{wrf}^{M4} (t) & RH_{wrf}^{M4} (t) & GHI_{wrf}^{M4} (t) & DHI_{wrf}^{M4} (t) & DI_{wrf}^{M4} (t)\end{pmatrix}
\end{align}

\begin{figure}[htbp]
\begin{center}
\centerline{\includegraphics[width=0.5\textwidth]{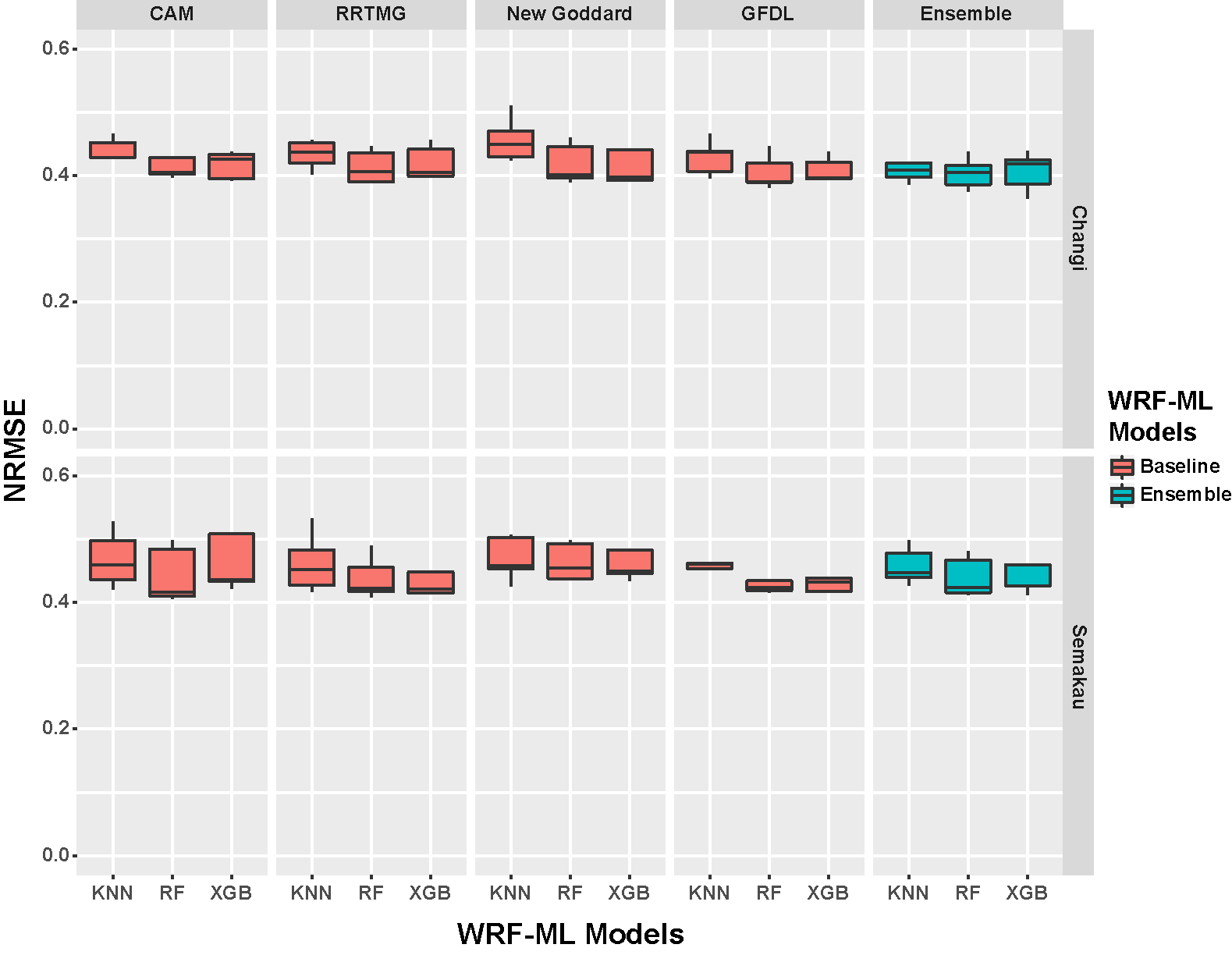}}
\caption{Plots of the normalized root-mean square error (NRMSE) for WRF-ML models with either WRF outputs for a single radiation model or an ensemble of 4 WRF outputs from 4 radiation models}
\label{fig:four-models}
\end{center}
\end{figure}

While others have shown that an ensemble of ML models can improve the predictive performance of the WRF-ML model, results as plotted in Fig.~\ref{fig:four-models} indicate that the predictive performance of an ensemble of four radiation models did not improve when compared to the use of single radiation models. 

Hence, this suggests that while the systematic biases for each radiation model can be individually compensated for, the current choice of predictive features are insufficient for the ML models to further differentiate circumstances under which any of the individual WRF radiation models might be more accurate. This also ties in with our intuition that frequently, a lot of the improved radiation (or other physics) models are developed to reduce systematic biases in the entire multi-physics model. Hence, the addition of a statistical algorithm on the output essentially reduces the systematic biases across all the models to similar levels. As ensembling typically improves performance by reducing variance across models, this also potentially explains why an ensemble of radiation models produces little improvement.

For completeness, the addition of actual measurement data as an additional feature vector was also tested and compared to the corresponding models that incorporated meteorological data as per Eq.~\ref{eq:t_model_24}. 

\begin{align} \label{eq:four_models_24}
   GHI_{pred} (t) = f \begin{pmatrix} & & &t & & \\
	                        &T_{wrf}^{M1} (t) & RH_{wrf}^{M1} (t) & GHI_{wrf}^{M1} (t) & DHI_{wrf}^{M1} (t) & DI_{wrf}^{M1} (t) \\
													&T_{wrf}^{M2} (t) & RH_{wrf}^{M2} (t) & GHI_{wrf}^{M2} (t) & DHI_{wrf}^{M2} (t) & DI_{wrf}^{M2} (t) \\
													&T_{wrf}^{M3} (t) & RH_{wrf}^{M3} (t) & GHI_{wrf}^{M3} (t) & DHI_{wrf}^{M3} (t) & DI_{wrf}^{M3} (t) \\
													&T_{wrf}^{M4} (t) & RH_{wrf}^{M4} (t) & GHI_{wrf}^{M4} (t) & DHI_{wrf}^{M4} (t) & DI_{wrf}^{M4} (t) \\
													&T_{met} (t - 24) & RH_{wrf} (t - 24) & GHI_{met} (t - 24) & DHI_{met} (t - 24) & DI_{met} (t - 24) \end{pmatrix}
\end{align}

\begin{figure}[htbp]
\begin{center}
\centerline{\includegraphics[width=0.5\textwidth]{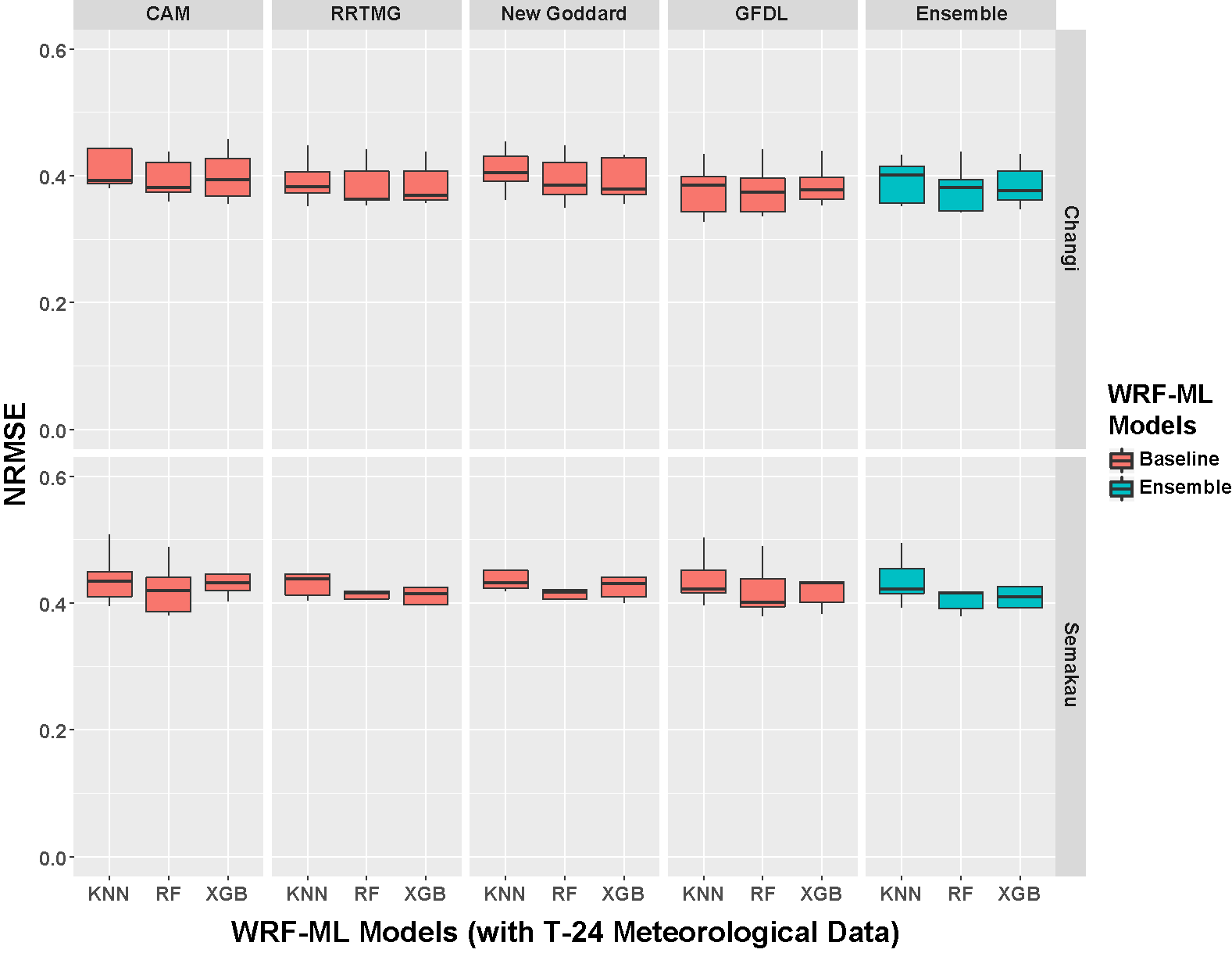}}
\caption{Plots of the normalized root-mean square error (NRMSE) for WRF-ML models with either WRF outputs for a single radiation model or an ensemble of 4 WRF outputs from 4 radiation models and additional meteorological data}
\label{fig:four-models-T24}
\end{center}
\end{figure}

Similarly, while an improvement over the simple ensemble model as described in Eq.~\ref{eq:four_models_base} was observed, there does not appear to be a significant improvement in predictive performance by using a combination of all 4 radiation model outputs. The inputs for the model are described in Eq.~\ref{eq:four_models_24} while computed NRMSEs are plotted in Fig.~\ref{fig:four-models-T24}.

\section{Discussion} \label{Section:4}

In this work, we investigated the use of different radiation models within WRF for the day-ahead forecasting of solar irradiance at 2 locations in the tropical city of Singapore. The NRSME of the two locations differ slightly, with Semakau island generally yielding slightly higher errors. Semakau island is expected to have slightly greater errors as the land area and land usage pattern near Semakau island as provided to WRF experience more substantial changes from new construction and land reclamation projects. While the NRMSE of the different models can differ slightly, we note in our experiments across both locations in Singapore that the general observations are fairly consistent, especially with regards to the improvements observed via the creation of hybrid WRF-ML models.  

In general, it is expected that the different WRF models will have some systemic biases in their predictions due to fundamental assumptions or simplifications in their derivation or have region-specific parameterization choices in the models. Hence, the use of ML models as an error correction mechanism for individual WRF models was proposed and indeed produced an improvement by an average of 20\% in the predictions in our test sets. Our experiments showed that K-NN models generally under-perform the two tree-based methods tested, i.e. gradient boosting and random forest, with the latter two models being fairly similar in performance. This could be due to the inherent simplicity of K-NN models relative to tree-based models. Hence, our preliminary results also suggest that the investigation and use of other more sophisticated statistical approaches may yield even larger improvements to the forecasting models.

In addition, the systemic biases in each radiation model are expected to be slightly different in magnitude based on their specific derivations, and we indeed observe that the newer RRTMG radiation model can out-perform the other radiation models without any error correction. However, it is interesting that the ML models, upon being applied to the WRF outputs, essentially improves each radiation model such that they have equivalent predictive performance after error correction. This also suggests that the methods outlined here could be equally applicable to improving biases in other WRF physics parameterizations, as long as one uses a consistent set of models as a baseline physics model. Critically, this also circumvents the typical lengthy evaluation of different WRF parameterizations that would be a necessity for forecasting based off purely NWP. While this work focused on solar forecasting, the proposed method can also be extended for use by wind farms via the training of a corresponding statistical model for wind speed and direction, and a possible extension to this work will involve further evaluation and validation for wind forecasting.

While many have postulated that ensemble models can improve the accuracy of weak learners, we did not notice a significant improvement in predictive performance when an ensemble of WRF outputs from all 4 radiation models are used as inputs for the ML models. This could potentially be due to the lack of additional predictive features that could be correlated to the better-performing radiation model. Ultimately, we note that the ensemble ML model only matched the performance of the individual hybrid WRF-ML models. This also suggests that the larger systemic biases in each radiation model have been accounted for, and that further features are required in order for the model to better learn the situations under which individual WRF radiation model predictions might be more accurate. The evaluation of this method with the inclusion of more WRF outputs as ML model features is certainly worth studying in subsequent work.

Lastly, we tested a new paradigm of utilizing actual meteorological data as input features in our hybrid ML error correction approach, and demonstrate that this can also bring about some benefit in further reducing the error. While the effect is naturally larger with more recent measured data, we note that the use of day-before data (t-24 time point measurements) can still improve the model's predictive performance. This concept could potentially be incorporated and further investigated, especially since many solar power plants typically also have weather stations co-located with them, which will serve as a high-quality source of data.

\section{Conclusion} \label{Section:5}

Solar and wind energy forecasting is critical for the incorporation of renewable energy in practical operations. While NWP methods can be accurate, the choice of physics models can require significant domain knowledge and insight. We show here that the use of a simple ML error correction model to post-process NWP predictions can significantly improve NRMSE in the predictions. More importantly, our results also suggest that the reduction in systematic errors across the different radiation models reduces the difference across these models, potentially removing the need for specific model tuning. Specifically, this is a unique outcome of the current work that has not been demonstrated or discussed in prior work using hybrid forecasting to the authors' knowledge. Lastly, we also propose and demonstrate that other models incorporating met mast sensor data can be beneficial for improving predictive accuracy, and propose further evaluation of such models in future work.



\bibliography{mybibfile}

\end{document}